\documentclass{article}
\usepackage{amsmath, amssymb}

\usepackage[hmargin=2.54cm, vmargin=3.18cm]{geometry}
\usepackage{graphicx}

\usepackage{hyperref}
\begin{document}
\title{A Knowledge-Enhanced Disease Diagnosis Method Based on Prompt Learning and BERT Integration}
\author{Zhang Zheng  Wu Hengyang}
\date{(Shanghai Polytechnic University, School of computer and Information Engineering, Shanghai 201209)}
\maketitle

\begin{abstract}
    This paper proposes a knowledge-enhanced disease diagnosis method based on a prompt learning framework. The method retrieves structured knowledge from external knowledge graphs related to clinical cases, encodes it, and injects it into the prompt templates to enhance the language model's understanding and reasoning capabilities for the task. We conducted experiments on three public datasets: CHIP-CTC, IMCS-V2-NER, and KUAKE-QTR. The results show that the proposed method significantly outperforms existing models across multiple evaluation metrics, with an F1 score improvement of 2.4\% on the CHIP-CTC dataset, 3.1\% on the IMCS-V2-NER dataset, and 4.2\% on the KUAKE-QTR dataset. Additionally, ablation studies confirmed the critical role of the knowledge injection module, as the removal of this module resulted in a significant drop in F1 score. The experimental results demonstrate that the proposed method not only effectively improves the accuracy of disease diagnosis but also enhances the interpretability of the predictions, providing more reliable support and evidence for clinical diagnosis.
\end{abstract}


\section{Introduction}
Disease diagnosis is the process of systematically identifying and confirming a patient's illness. It involves analyzing the patient's symptoms, medical history, physical examination results, and data obtained from various medical tests, such as laboratory tests and imaging studies. The purpose of disease diagnosis is to determine the cause of the illness, thereby guiding physicians in selecting appropriate treatment methods and increasing the patient's chances of recovery. Disease diagnosis holds critical importance in medicine. It serves as the foundation for formulating treatment plans, allowing doctors to prescribe targeted treatment regimens and avoiding misdiagnosis and over-medication. Moreover, accurate diagnosis helps prevent further deterioration and spread of diseases, particularly in the control of infectious diseases. On the other hand, disease diagnosis can enhance the efficiency of medical resource utilization, reduce unnecessary tests and treatments, and lower healthcare costs.

Initially, people primarily relied on doctors' experience and knowledge for diagnosis, using methods such as patient interviews, physical examinations, and laboratory tests. However, this simple and mechanical approach clearly could not produce optimal diagnostic results. To address issues of subjectivity, time consumption, and risks of misdiagnosis or missed diagnosis, knowledge engineering methods were later adopted\cite{ref1}. These methods utilize rule-based matching techniques to determine text categories.However, in recent years, with the rapid development of technologies such as big data and cloud computing, the internet has experienced an explosive growth in information. The vast amount of textual data has posed significant challenges to traditional classification methods. In response to this, efforts have been made to develop new algorithms and models that integrate multi-source information, including clinical data, medical imaging, and biomarkers, to enhance the accuracy and reliability of diagnoses.For example, Luo et al.\cite{ref2} proposed a new multi-modal heterogeneous graph model to represent medical data, which effectively addresses the challenge of label allocation within the same cluster. This model enables more precise diagnostic targeting, improving the accuracy of diagnosis.

In addition, with the rapid advancement of artificial intelligence and natural language processing technologies, the use of deep learning models for text classification to assist in disease diagnosis has become a significant area of research.Among them, the BERT model, as a powerful natural language processing tool, has demonstrated outstanding performance in many text classification tasks.Onan et al.\cite{ref3}proposed an innovative hierarchical graph-based text classification framework, which captures the complex relationships between nodes through a dynamic fusion mechanism of contextual node embeddings and the BERT model.Although these methods have shown good performance in terms of classification accuracy and stability, they still face many challenges in practical applications. The main issues are related to poor structured reasoning and low efficiency in medical data annotation.

To address this issue, this paper proposes an integrated approach that combines pre-trained language models with knowledge graphs for extracting and processing structured knowledge from clinical texts, which is then applied to disease prediction tasks. Experimental results show that the proposed method achieves superior performance on public datasets for disease diagnosis.In addition, experiments have demonstrated that knowledge injection based on prompt learning effectively guides language models in acquiring relevant medical knowledge, thereby enhancing reasoning performance. Through this method, we achieved the conversion from raw text to high-quality semantic vectors, and mapped these vector representations to specific vocabulary using a function.This method not only enhances the semantic representation capabilities of the text but also provides rich contextual information and structured knowledge support for disease prediction, thereby improving the model's performance and reliability in practical applications. The following sections of this paper will provide a detailed explanation of the implementation steps and experimental results.

\section{Related work}
The continuous advancement of information technology has driven the development of intelligent clinical decision support systems. The widespread application of machine learning models has significantly improved the effectiveness of these systems, particularly in the field of disease prediction. The evolution of these models has gone through several stages: from the early expert rule-based models, to models based on statistical analysis and case-based reasoning, and finally to the current advanced models utilizing machine learning and deep learning techniques\cite{ref4}.At the same time, breakthroughs in the field of natural language processing (NLP) have introduced innovative tools and techniques, opening up new perspectives for disease diagnosis and providing unprecedented possibilities.

\subsection{Disease Diagnosis}
In these early disease diagnosis methods, rule-based research primarily relied on the analysis of medical literature and case data. Expert rule-based disease diagnosis methods involved the collection of expert diagnostic experiences to form disease diagnosis pathways, thus creating expert systems. A typical example of such an expert system is the MYCIN expert system\cite{ref5}, developed by Shortliffe in 1976, which became a foundational model for many subsequent expert systems in the medical field.However, many techniques are unable to explain the processes involved in disease monitoring and data inference. To address this, Aami and Sarfraz\cite{ref6} proposed a fuzzy rule-based diabetes classification system, which combines fuzzy logic with the cosine amplitude method. They developed two fuzzy classifiers, and the proposed model demonstrated high predictive accuracy.Sanz et al. \cite{ref7}further developed a new approach based on the Fuzzy Rule-Based Classification System (FRBCS) by integrating it with Interval-Valued Fuzzy Sets (IV-FRBCS). They validated the applicability and effectiveness of this method in medical diagnostic classification problems, demonstrating its potential in improving diagnostic accuracy in complex medical scenarios.With the accumulation of large-scale clinical data, statistical analysis has become an important method for disease diagnosis. Researchers use statistical analysis to uncover potential correlations between patient characteristics and medical indicators, thereby providing new perspectives and approaches for disease prediction and diagnosis. As a result, statistically based disease diagnosis methods have become increasingly significant in medical research and practice.Lv et al. \cite{ref8}developed a multi-classification and disease diagnosis model based on methods such as multinational logistic regression and discriminant analysis, focusing on factors like sleep quality. This classification model can be utilized during the early stages of disease diagnosis to categorize different diseases, aiding in the initial diagnostic process.Yadav et al.\cite{ref9}discussed statistical modeling and prediction techniques for various infectious diseases, addressing the issue of single-variable data related to infectious diseases. They proposed fitting time series models and making predictions based on the best-fit model, offering a more accurate approach to forecasting the spread and development of infectious diseases.

Recently, neural network-based disease diagnosis methods have emerged and gradually become a hot topic of research. Researchers are focused on developing new algorithms and models that integrate multi-source information, such as clinical data, medical imaging, and biomarkers, to enhance the accuracy and reliability of diagnoses. These advancements aim to improve diagnostic capabilities by leveraging the power of deep learning in processing complex medical data.Wang and Li employed machine learning methods, combined with large-scale clinical databases, to develop a statistical-based disease prediction model. This model successfully achieved diagnostic predictions for multiple diseases, offering new insights into personalized disease diagnosis. Their approach represents a significant advancement in tailoring diagnoses to individual patient data.On the other hand, to address the challenge of assigning specific stages in the diagnosis of clinical diseases with long courses and staging characteristics, Ma et al.\cite{ref10} proposed a gout staging diagnosis method based on deep reinforcement learning. They first used a candidate binary classification model library to accurately diagnose gout, and then refined the binary classification results by applying predefined medical rules for gout diagnosis. Additionally, some scholars have integrated diagnostic rules into deep neural networks for bladder cancer staging, significantly mitigating the drawback of cancer staging methods based on deep convolutional neural networks that tend to overlook clinicians' domain knowledge and expertise.Additionally, to address the impact of imbalanced medical record samples on the training and predictive performance of disease prediction models, the academic community has proposed various solutions for training models on small and imbalanced datasets [11-12]. These methods aim to improve model robustness and accuracy when faced with limited or skewed data distributions, which is a common challenge in medical data analysis.

However, despite the numerous advantages brought by neural network-based disease diagnosis methods, there are still some shortcomings. First, the aforementioned methods often require large amounts of labeled data for training, and annotating medical data typically demands significant time and effort from expert physicians, making it costly and time-consuming. Second, due to the complexity and diversity of medical data, existing machine learning models may struggle to adapt well to various diseases and clinical scenarios, leading to insufficient generalization capability of the models.Through further reading and research, it was found that many scholars have proposed different improvement methods to address this challenge.For example, Luo et al.\cite{ref13} first proposed a new multi-modal heterogeneous graph model to represent medical data, which helps address the label allocation challenges within the same cluster, enabling more precise targeting of desired medical information.At the same time, R.D. et al. proposed a domain knowledge-enhanced multi-label classification (DKEC) method for electronic health records, addressing the issue of previous work neglecting the incorporation of domain knowledge from medical guidelines. They introduced a label attention mechanism and a simple yet effective group training method based on label similarity\cite{ref14}.This method greatly improves applicability to minority (tail) class label distributions.

\subsection{Application of the BERT Model in Disease Diagnosis}
BERT (Bidirectional Encoder Representations from Transformers), proposed by Google in 2018, is an advanced pre-trained natural language processing (NLP) model. This model has achieved state-of-the-art results in various NLP tasks, including but not limited to text classification, named entity recognition (NER), and question-answering systems (QA).Recently, the research community has begun exploring the potential of the BERT model in the medical field. By integrating multi-source information such as clinical texts and medical literature, the BERT model can absorb and learn rich medical knowledge, thereby providing more precise auxiliary information in the disease diagnosis process. The bidirectional contextual modeling capability of the BERT model gives it a significant advantage in understanding the complex contexts of medical texts, offering comprehensive informational support for disease diagnosis.

In the research on medical named entity recognition algorithms, Tian et al. \cite{ref15}first proposed a method based on pre-trained language models. This method utilizes the BERT model to generate sentence-level feature vector representations of short text data and combines it with recurrent neural networks (RNN) and transfer learning models to classify medical short texts.Onan et al.\cite{ref16} proposed an innovative hierarchical graph-based text classification framework to enhance the performance of text classification tasks. This framework effectively captures the complex relationships between nodes in the hierarchical graph through a dynamic fusion mechanism of contextual node embeddings and the BERT model.Xu et al.\cite{ref17} developed a medical text classification model (CMNN) that combines BERT, convolutional neural networks (CNN), and bidirectional long short-term memory networks (BiLSTM) to address the efficiency and accuracy challenges in medical text classification. This model showed improvements in evaluation metrics such as accuracy, precision, recall, and F1 score compared to traditional deep learning models.

As advanced NLP technologies gradually penetrate the medical field, the BioBERT model has emerged. This model is specifically optimized for the biomedical domain, demonstrating significant advantages in the understanding and processing of medical texts.Sharaf et al.\cite{ref18}provided a detailed analysis and overview of a systematic BioBERT fine-tuning method aimed at meeting the unique needs of the healthcare field. This approach includes annotating data for medical entity recognition and classification tasks, as well as applying specialized preprocessing techniques to handle the complexity of biomedical texts.

Recently, pre-trained language models have achieved significant success in many question-answering tasks[19-20]. However, while these models encompass a wide range of knowledge, they perform poorly in structured reasoning. Additionally, considering the previously mentioned issues, such as the low efficiency of medical data annotation, these limitations pose further challenges.This paper aims to enhance structured reasoning using knowledge graphs and leverage the BERT model's outstanding performance in text classification tasks. The primary focus of this research is on how to effectively utilize language models and knowledge graphs for reasoning, thereby achieving the goal of significantly improving disease prediction accuracy.

\section{Task formulation}
Given clinical information and electronic medical records of a patient, where $w_i$ represents the $i$-th word in the text and $N$ represents the total number of words, this research aims to predict the disease type $\hat{y}$ that the patient has based on the content of $x$.Therefore, this task can be represented as learning a model $f(\cdot)$ with parameters $\theta$.Given an input $x, f(x ; \theta)$ outputs a predicted result $\hat{y} \in \mathcal{Y}$. Here, $\mathcal{Y}$ is the set of labels for all candidate disease types.

This paper utilizes prompt learning to accomplish the task, converting it from a classification problem into a language modeling problem. The original classification problem is formulated as fitting $Y=P(X ; \theta)$. The transformed language modeling problem is formulated as fitting $Y=(T(X) ; \theta)$, where $T(X)$ represents the prompt template used to encapsulate the original text into a new input.

\section{Methodology}
\begin{figure}[h]
    \centering
    \includegraphics[width = .65\textwidth]{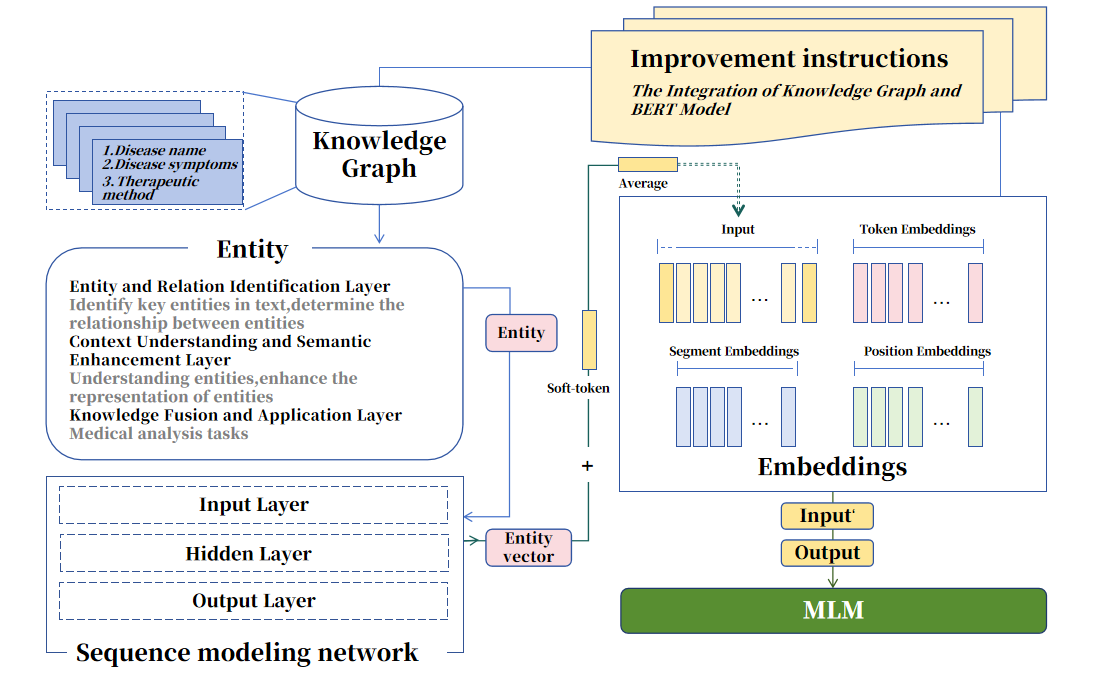}
    \caption{Method Structure Diagram}
\end{figure}

\subsection{Knowledge Retrieval}
Clinical texts contain many conceptual and structured forms of knowledge, such as symptoms, diagnoses, and treatment plans. Knowledge retrieval is capable of identifying and extracting key medical concepts and relationships.For example, ``Type 2 diabetes" is a common chronic disease, characterized primarily by persistently elevated blood glucose levels. The diagnostic criteria include fasting blood glucose and HbA1c levels. Common symptoms include excessive thirst, hunger, frequent urination, weight loss, and fatigue.Through knowledge retrieval, it can extract structured knowledge such as causes, related complications, risk factors, and more. This type of knowledge plays an important auxiliary role in disease prediction. Therefore, this paper utilizes the following methods to achieve knowledge retrieval from the text:

Given the clinical text $x$ of a patient, we process $x$ by performing tasks such as tokenization and part-of-speech tagging, resulting in a set of vocabulary $\mathrm{x}=\left\{\mathrm{w}_1, \mathrm{w}_2, \ldots, \mathrm{w}_{\mathrm{n}}\right\}$. First, we apply the named entity recognition (NER) model $f_{N E R}$ to identify the set of entities $E_x=\left\{e_1, e_2, \ldots e_m\right\}$.
\begin{equation}
    \mathrm{E}_{\mathrm{x}} \rightarrow \mathrm{f}_{\mathrm{NER}}(\mathrm{x})
\end{equation}

Let the knowledge graph be $\mathcal{G}=(V, E)$, where the entity set is $\mathrm{V}=\left\{\mathrm{v}_1, \mathrm{v}_2, . . \mathrm{v}_{\mathrm{k}}\right\}$. V may include entities such as ``Type 2 diabetes," ``hyperglycemia," ``insulin," ``weight gain," etc. The edge set E may include relationships such as ``is symptom," ``used for treatment," ``causes," and others.Next, for each entity $e_i$ extracted from the text, we find the corresponding entity $v_i$ in the knowledge graph. The key to this matching process is determining whether $e_i$ and $v_i$ represent the same or similar concepts. This process can be implemented using a similarity function $\operatorname{sim}=\left(e_i, v_i\right)$. The entities $e_i$ and $v_i$ are considered a match when $\operatorname{sim}=\left(\mathrm{e}_{\mathrm{i}}, \mathrm{v}_{\mathrm{i}}\right)$ reaches a certain threshold $\theta$. When the conditions $\forall e_i \in E_x, \exists v_i \in G_E$ and $\forall e_i \in E_x, \exists v_i \in G_E: \operatorname{sim}\left(e_i, v_i\right) \geq \theta$ are satisfied, they are considered to represent the same or related entities.

Next, for each pair of matched entities, we search for all possible reasoning paths $P$ in the knowledge graph. Relationships in the knowledge graph can be represented in the form of triples $\left(\mathrm{v}_{\mathrm{i}}, \mathrm{r}, \mathrm{v}_{\mathrm{j}}\right)$, where $v_i, v_j \in G$ are entities, and $r \in G_R$ is the relationship. The goal of inference is to find possible paths P from entity $v_i$ to $v_j$.
\begin{equation}
    \mathrm{P}\left(v_i, v\right)=\left\{\left(v_i, r_1, v_{i 1}\right),\left(v_{i 1}, r_2, v_i\right), \ldots,\left(v_{i n}, r_n, v_j\right)\right\}
\end{equation}

For each pair of matched entities $\left(v_i, v_j\right)$, collect all possible reasoning paths $P\left(v_i, v_j\right)$, thereby obtaining structured knowledge between entities extracted from clinical texts. This knowledge is then utilized for further analysis and applications, such as disease prediction and relationship inference.For example, the entity extracted from the text $e_t=$ ``type 2 diabetes", and the next step is to find the matching entity in the knowledge graph $v_t=$ ``Type 2 Diabetes". If the similarity function $\operatorname{sim}\left(e_t, v_k\right) \geq \theta$, they are considered a match.Finding the path P: (``Type 2 Diabetes", ``causes", ``High Blood Sugar") $\rightarrow$ (``High Blood Sugar", ``leads to", ``Kidney Disease"). Next, collect and record the path, constructing the reasoning chain from ``Type 2 Diabetes" to ``Kidney Disease".
\begin{equation}
    \operatorname{Paths}\left(\mathrm{v}_{\mathrm{i}}, \mathrm{v}_{\mathrm{j}}\right)=\mathrm{U}_{\text {all valid }} \mathrm{P}\left(\mathrm{v}_{\mathrm{i}}, \mathrm{v}_{\mathrm{j}}\right)
\end{equation}

Next, we will represent the structured knowledge along these reasoning paths.

\subsection{Knowledge Representation}
The reasoning paths from the collection are concatenated into a single text sequence. This concatenated text is then represented as a vector $k$ using a model.

(1) Reasoning Path Representation

Each reasoning path $\mathrm{P}\left(\mathrm{g}_{\mathrm{i}}, \mathrm{g}_{\mathrm{j}}\right)$ is converted into a readable text sequence, with each path consisting of a series of triples $\left(\mathrm{g}_{\mathrm{i}}, \mathrm{r}, \mathrm{g}_{\mathrm{j}}\right)$. These triples can be transformed into text, where each triple is expressed in the form of ``Entity 1 reaches Entity 2 through Relationship."For example, given a reasoning path $\left(\mathrm{g}_1, \mathrm{r}_1, \mathrm{g}_2\right),\left(\mathrm{g}_2, \mathrm{r}_2, \mathrm{g}_3\right)$, it can be converted into the text: ``Entity $g_1$ reaches Entity $g_2$ through Relationship $r_1$, and $\mathrm{g}_2$ reaches Entity $g_3$ through Relationship $r_2$. "Assuming the knowledge graph contains the following triples: (``Type 2 Diabetes", ``causes", ``High Blood Sugar"), (``High Blood Sugar" , ``leads to" , ``Kidney Disease"), the conversion to text would be: ``Type 2 Diabetes reaches High Blood Sugar through causes, and reaches Kidney Disease through leads to."

After converting the path into text, the pre-trained BERT model $M$ is used to transform the text $T_p$ into a vector representation $k_p$.

(2) Text Vectorization Model Selection (BERT)

The input text $T_p$ is first tokenized by BERT, splitting the input text into smaller subword units. For example, the input text ``Type 2 diabetes passed causes to high blood sugar, pass leads to Kidney disease" might be tokenized into subwords like ``Type 2", ``diabetes", ``pass", ``causes", ``to", ``Kidney disease", ``to", and so on.Subsequently, each token is converted into an ID from BERT's vocabulary. After tokenization, the next step involves encoding, where the token ID sequence is transformed into embedding vectors. These embedding vectors are then fed into BERT's multi-layer Transformer encoder, effectively capturing the contextual relationships and dependencies between the tokens. The formula for this process is as follows:
\begin{equation}
    \mathrm{f}_{\text {BERT }}: \mathrm{x}_{\mathrm{p}}=\mathrm{f}_{\text {BERT }\left(\mathrm{T}_{\mathrm{P}}\right)}
\end{equation}
Where:

$T_p$ is the input text sequence.

$x_p$ is the text vector representation output by the BERT model.

The function $f_{B E R T}$ transforms the text sequence $T_p$ into its vector representation $x_p$. Here, $x_p$ contains the semantic information of the input text $T_p$.

The BERT model outputs the vector representation of the text $T_p$ as:
\begin{equation}
    k_p=M\left(x_p\right)
\end{equation}

Here, $x_p$ is the vector representation obtained by processing the input text sequence $T_p$ through the BERT model. It captures the semantic information of the input text.M is an additional mapping function that takes $x_p$ as input and further processes it. $k_p$ is the final text vector representation obtained after processing by the mapping function M.

(3) Output Vector Extraction

In the text vectorization model selection mentioned earlier, we used the BERT model to convert the input text sequence $T_p$ into the vector representation $x_p$. During this process, we typically choose to use the special token[CLS], and its corresponding vector is used as the overall representation of the entire input text. This comprehensive representation is denoted as $k_p[0]$.

\subsection{Prompt Template}
(1) Construction of Prompt Template

To effectively utilize the BERT model for text vectorization, we first need to preprocess the original input text x to construct a proper prompt template Template $(x)$ that meets the input format requirements of the BERT model.This prompt template will generate a text $x'$ that contains special tokens and specific vocabulary.The text $x'$ contains special tokens such as [MASK], soft tokens, and other relevant vocabulary. These elements help the model focus on specific parts of the input and guide the prediction process, enhancing the performance of tasks like classification or inference.

For example:

Original input text $x$: ``Type 2 diabetes is a chronic metabolic disease."

Prompt template Template $(x)$: ``Type 2 diabetes is a[MASK]type[SOFT]disease."

Generated preprocessed text $x'$: ``Type 2 diabetes is a[MASK]type[SOFT]disease."

(2)Expression of Template Conversion
\begin{equation}
    x^{\prime}=\text { Template }(\mathrm{x})=\left[\mathrm{w}_1, \text { soft, } \mathrm{w}_2, \text { mask }, \mathrm{w}_3, \ldots\right]
\end{equation}

Here, $w_i$ represents either words from the original text or other words added based on the context. Suppose we want to convert the original text x into the template $x^{\prime}$. This can be achieved by inserting special tokens and soft labels. A specific example is as follows:

Original input text $x$: ``Type 2 diabetes is a chronic metabolic disease."

Converted preprocessed text $x'$: ``Type 2 diabetes is a[MASK]type[SOFT]disease."

It can be represented as:
\begin{equation*}
    x^{\prime}=\text { Template }(x)=[\text {``Type 2 diabetes"}, soft, \text{``is a"}, mask, \text{``type"}, soft, \text{``disease".}] 
\end{equation*}

(3) Vector Representation and Processing

Using the pre-trained model M , the preprocessed text $x^{\prime}$ is transformed into a set of vectors $\left\{k_1, k_{\text {soft }}, k_2, k_{\text {mask }}, k_3, \ldots\right\}$. Each vector $k_i$ corresponds to a word or token in the preprocessed text $x^{\prime}$, where:

$k_{\text {soft }}$ is the vector corresponding to the soft token.

$k_{\text {mask }}$ is the vector corresponding to the mask token.

For example, for the original input text $x$: ``Type 2 diabetes is a chronic metabolic disease.", the generated vector set can be explained as follows:

$k_1$ : The vector corresponding to the first word in the original text, ``Type 2 diabetes". 

$k_{\text {soft }}$ : The vector corresponding to the soft token.

$k_2$ : The vector corresponding to the second word in the original text, ``is a". 

$k_{\text {mask }}$ : The vector corresponding to the mask token.

$k_3$ : The vector corresponding to the third word in the original text, ``disease".

Other vectors follow similarly, with each vector representing the respective words or tokens in the preprocessed text.

(4) Vector Processing

The formula for replacing the original $k_{\text {soft }}$ vector with the average of $k_{\text {soft }}$ and other related vectors $k_i$ is as follows:
\begin{equation}
    k_{\text {soft }}^{\text {new }}=\frac{1}{n+1}\left(k_{\text {soft }}+\sum_{i=1}^n k_i\right)
\end{equation}
Where:
$k_{\text {soft }}^{\text {new }}$ is the updated $k_{\text {soft }}$ vector,

$k_{\text {soft }}$ is the original soft token vector,

$k_i$ represents related vectors (e.g., corresponding to related tokens or words), 

$n$ is the number of related vectors used in the calculation.

Finally, after replacing the original $k_{\text {soft }}$ with $k_{\text {soft }}^{\text {new }}$, we obtain the updated vector set: $\left\{k_1, k_{s o f t}^{\text {new }}, k_2, k_{\text {mask }}, k_3, \ldots\right\}$, which will be used in the next steps of the model processing.

\subsection{Prediction}
The text $x^{\prime}$ containing the mask and the new soft token is input into a pre-trained language model for forward inference. This produces representations for each token. The representation at the mask position is then used for a verbalizer prediction.

Given the modified input $x^{\prime}$, it is formalized as:
\begin{equation}
    x^{\prime}=\left[\mathrm{w}_1, \text { softnew }, \mathrm{w}_2, \text { mask }, \mathrm{w}_3, \ldots\right]
\end{equation}

Feed $x^{\prime}$ into the pre-trained language model $M$, and perform forward inference to obtain the vector representations for each token:
\begin{equation}
    \left\{\mathrm{k}_1, \mathrm{k}_{\mathrm{soft}}^{\text {new }}, \mathrm{k}_2, \mathrm{k}_{\text {mask }}, \mathrm{k}_3, \ldots\right\}
\end{equation}

Extract the representation $k_{\text {mask }}$ of the mask token from the model output.

Use a verbalizer function $V$ to map $k_{\text {mask }}$ to a specific vocabulary term. The Verbalizer $V$ is a mapping from the vector space to the vocabulary, typically used to convert the model's predicted vector into a specific word.
\begin{equation}
    \text { predicted word }=\mathrm{V}\left(\mathrm{k}_{\text {mask }}\right)
\end{equation}

The specific steps are as follows:

(1) Calculate the similarity between $k_{\text {mask }}$ and the embedding vectors of each word in the vocabulary to obtain a probability distribution. For example, if the embedding vector for each word in the vocabulary is $e_i$, then the probability distribution is given by:
\begin{equation}
    \text { similarity }=\mathrm{k}_{\text {mask }} \cdot \mathrm{e}_{\mathrm{i}}
\end{equation}

(2) Based on the computed similarities, select the word with the highest probability as the prediction result.

\section{Experimental settings}
\subsection{Datasets}
In this paper, experiments were conducted on the CHIP-CTC, IMCS-V2-NER, and KUAKE-QTR datasets. CHIP-CTC, one of the experimental datasets, originates from a bench-marking task released at the CHIP2019 conference. All text data is sourced from real clinical trials, including 22,962 entries in the training set, 7,682 entries in the validation set, and 10,000 entries in the test set. The dataset is available at \url{https://github.com/zonghui0228/chip2019task3.}

The IMCS-V2-NER dataset, used as another experimental dataset, comes from the Named Entity Recognition task in the IMCS2 dataset developed by the School of Data Science at Fudan University. It includes 2,472 entries in the training set, 833 entries in the validation set, and 811 entries in the test set. The dataset is available at \url{https://github.com/lemuria-wchen/imcs21.}

The KUAKE-QTR dataset, used as another experimental dataset, includes 24,174 entries in the training set, 2,913 entries in the validation set, and 5,465 entries in the test set. The dataset is available at \url{https://tianchi.aliyun.com/dataset/95414.}
\begin{table}[h]
    \centering
    \caption{Datasets}
    \begin{tabular}{lccc}
        \hline
        Number of samples / dataset	& CHIP-CTC	& IMCS-V2-NER &	KUAKE-QTR \\
        \hline
        Number of training samples	&22962	&2472&	24174\\
        Number of validation samples&	7682&	833	&2913\\
        Number of test samples&	10000&	811&	5465\\
        \hline
    \end{tabular}
\end{table}

\subsection{Baseline}
When evaluating our proposed method, we established a comprehensive set of baseline models to ensure rigorous and fair comparison. These baselines were selected to represent robust benchmarks in the field, including both traditional machine learning algorithms and advanced deep learning techniques.

SVM: Support Vector Machines (SVM) differentiate between classes by finding the optimal hyperplane, effectively handling both linearly separable and non-linearly separable problems. This classifier uses kernel techniques to process text data, enabling efficient identification and classification of complex text patterns in tasks such as sentiment analysis and spam detection.

CNN: Convolutional Neural Networks (CNNs) extract local features from text data using convolutional layers in text classification tasks. These local features capture key semantic information within sentences or documents, helping the model understand and classify the text content. This ability of CNNs is particularly well-suited for scenarios where quick and effective identification of key information from text is required.

RNN: Recurrent Neural Networks (RNNs) are particularly well-suited for text classification tasks due to their ability to process sequential text data and capture long-term dependencies between words. This allows RNNs to understand contextual information within the text, leading to more accurate predictions of text categories, such as sentiment or topic classification.

BiLSTM: Bidirectional Long Short-Term Memory (BiLSTM) networks combine both forward and backward LSTMs, allowing them to consider both preceding and succeeding contextual information in the text. This structure makes BiLSTMs particularly effective for text classification tasks as they can capture temporal dependencies in the text data comprehensively. By doing so, BiLSTMs provide a deeper semantic understanding, resulting in higher accuracy across various text classification scenarios.

Attention: In text classification tasks, the Attention mechanism enables the model to focus on key parts of the text, enhancing its understanding of the overall content's importance. By assigning different weights to various words or phrases, it emphasizes the information most influential for the classification task.

BiRNN: Bidirectional Recurrent Neural Networks (BiRNN) combine both forward and backward RNNs, allowing them to consider contextual information from both directions in a text sequence. This structure enables BiRNNs to better capture long-term dependencies in text data, thereby enhancing classification performance. It is particularly well-suited for tasks that require a comprehensive understanding of text semantics, such as sentiment analysis and topic classification.

BERT: BERT (Bidirectional Encoder Representations from Transformers) plays a crucial role in text classification tasks. By utilizing pre-trained models to understand the semantics and context of text, and through fine-tuning techniques, BERT can capture rich textual features and improve classification accuracy. Its bidirectional encoding and context-aware capabilities enable BERT to excel across various text classification scenarios.

The inclusion of these baselines aims to provide a clear reference point for evaluating the performance of our new method. By comparing with these established models, we aim to demonstrate the advancements of our approach relative to existing solutions. Bench-marking against these well-established models helps highlight scenarios where our method offers superior performance, thereby validating its effectiveness and efficiency in addressing the problem at hand.

\section{Results and Analysis}
This section presents our experimental results and provides an analysis of these results. To ensure the reliability of the experiments, we repeated each experiment three times and used the average values as the final results.

\subsection{Comparison Experiments}
\begin{table}[h]
    \centering
    \caption{Comparison Experiment Results}
    \begin{tabular}{lccccccccc}
        \hline
        & \multicolumn{3}{c}{KUAKE-QTR} & \multicolumn{3}{c}{IMCS-V2-NER} & \multicolumn{3}{c}{CHIP-CTC} \\
        \hline
        & P&	R	&F1	&P	&R	&F1	&P	&R	&F1 \\
        SVM        & 0.53    & 0.48    & 0.49    & 0.41    & 0.40    & 0.41    & 0.45    & 0.19    & 0.23    \\
        CNN        & 0.56    & 0.50    & 0.38    & 0.65    & 0.60    & 0.65    & 0.65    & 0.61    & 0.60    \\
        RNN        & 0.75    & 0.60    & 0.62    & 0.73    & 0.70    & 0.72    & 0.71    & 0.72    & 0.71    \\
        BiLSTM     & 0.85    & 0.80    & 0.82    & 0.76    & 0.82    & 0.82    & 0.85    & 0.78    & 0.74    \\
        Attention  & 0.67    & 0.50    & 0.55    & 0.69    & 0.65    & 0.67    & 0.67    & 0.80    & 0.65    \\
        BiRNN      & 0.82    & 0.78    & 0.79    & 0.74    & 0.72    & 0.74    & 0.74    & 0.73    & 0.74    \\
        BERT       & 0.94    & 0.90    & 0.94    & 0.96    & 0.95    & 0.96    & 0.92    & 0.91    & 0.92    \\
        Ours       & 0.96    & 0.92    & 0.94    & 0.96    & 0.94    & 0.95    & 0.94    & 0.91    & 0.93    \\
        \hline
    \end{tabular}
\end{table}

\begin{figure}[h]
    \centering
    \begin{minipage}{.48\textwidth}
        \includegraphics[width = \linewidth]{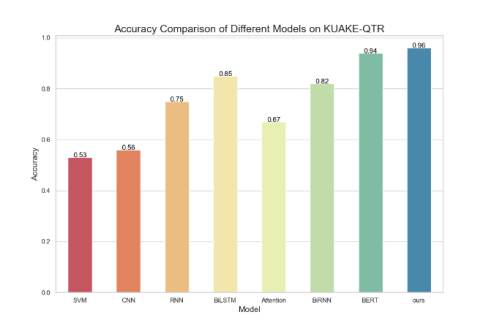}
        \caption{Performance of Different Benchmark Models and the Proposed Model on the KUAKE-QTR Dataset}
    \end{minipage}\hfill 
    \begin{minipage}{.48\textwidth}
        \includegraphics[width = \linewidth]{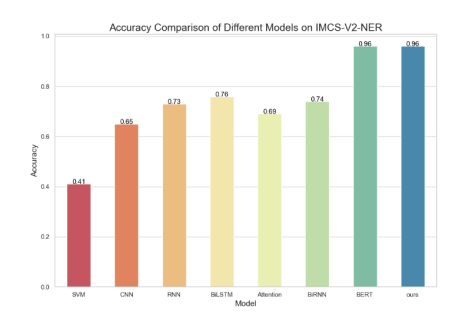}
        \caption{Performance of Different Benchmark Models and the Proposed Model on the IMCS-V2-NER Dataset}
    \end{minipage}

    \begin{minipage}{.48\textwidth}
        \includegraphics[width = \linewidth]{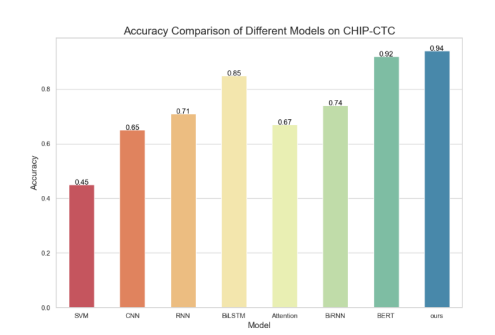}
        \caption{Performance of Different Benchmark Models and the Proposed Model on the CHIP-CTC Dataset}
    \end{minipage}\hfill 
    \begin{minipage}{.48\textwidth}
        \includegraphics[width = \linewidth]{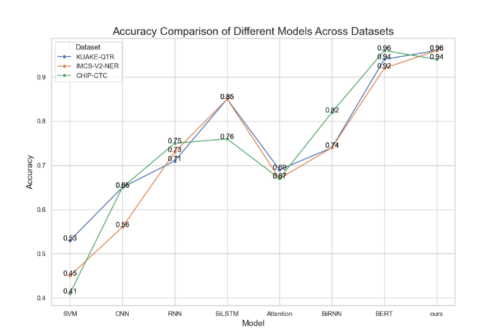}
        \caption{Accuracy Comparison of Eight Models Across Three Datasets}
    \end{minipage}
\end{figure}

The results indicate that as the complexity of the models increases, the accuracy improves accordingly across different datasets. SVM, as a classical machine learning model, can handle linearly separable data, but it struggles to capture more complex relationships when datasets have intricate features and patterns. CNN performs better than SVM on the CHIP-CTC, IMCS-V2-NER, and KUAKE-QTR datasets, as it excels at extracting local key information compared to SVM.The experimental results indicate that although the accuracy improves compared to SVM, the overall performance remains relatively low. CNNs show limited effectiveness in handling natural language tasks, particularly in capturing long-range dependencies. The Attention mechanism proves to be more effective at focusing on crucial parts of the input sequence, thereby enhancing the model's performance with sequential data. Compared to CNNs, Attention provides greater flexibility in highlighting key aspects of the data, resulting in improved accuracy. Additionally, BiRNNs exhibit significantly improved performance across various datasets. Unlike traditional RNNs, BiRNNs utilize two separate RNNs—forward RNN and backward RNN—to process the input sequence, which contributes to their superior performance.BiRNN addresses the limitation of traditional RNNs, which can only utilize past context and cannot leverage future context. By incorporating bidirectional processing, BiRNNs offer enhanced context understanding and feature extraction capabilities. The bidirectional feature extraction enables BiRNNs to better capture complex patterns and semantic relationships within sequences, thus improving the model's predictive performance.

In comparison, BiLSTM further enhances these capabilities by employing LSTM units with gating mechanisms that effectively mitigate the vanishing gradient problem. This allows BiLSTM to capture long-range dependencies and bidirectional context more effectively. Experimental results demonstrate that BiLSTM achieves higher accuracy than BiRNN, highlighting its advantages in handling complex sequence processing tasks.Compared to BiLSTM and other models, BERT leverages the self-attention mechanism in the Transformer architecture to simultaneously consider both left and right context information. Trained on extensive corpora and fine-tuned for specific tasks, BERT's multi-layer self-attention mechanism captures deeper and more nuanced features. This results in superior feature extraction capabilities compared to RNNs, BiRNNs, and BiLSTMs, making BERT more effective in understanding and processing complex textual data.

From the last row of the table, it is evident that our method achieved the best performance across all four datasets. This demonstrates the feasibility and effectiveness of our approach. Furthermore, it highlights the significant impact of integrating medical knowledge into the model through the prompt-learning framework, which notably enhances model performance.

\subsection{Ablation Study}
This section aims to analyze the effectiveness of each module in our proposed method. To achieve this, we first remove the knowledge representation component from our method.As seen in the first row of Table 3, removing the knowledge representation results in a significant drop of 0.2 in the F1 score. This is because knowledge representation is a core component of our method, which injects structured knowledge obtained from the knowledge graph into the language model, thereby enhancing the model's understanding of medical domain knowledge and improving performance on medical diagnostic tasks.

Subsequently, we modified the prompt template, changing it to: ``The characteristics of Type 2 Diabetes include [MASK] nature, and it is also manifested as [SOFT] disease."From rows 2 to 4 of Table 3, it can be observed that the prompt template used in our method achieves the best performance, while other prompt templates do not perform as well. We believe this may be due to the presence of excessive or irrelevant words in the prompt templates, which could introduce noise into the language model's reasoning process.

Finally, we experimented with the impact of different prediction words on the model's inference results. As shown in the last three rows of the table, inappropriate prediction words can severely mislead the model's reasoning, leading to a significant decrease in the final disease diagnosis results.
\begin{table}[h]
    \centering
    \caption{Ablation Experiment Results}
    \begin{tabular}{lccc}
        \hline
        &P	&R	&F1 \\
        \hline
        Our method	&0.93	&0.92	&0.91\\
        Remove knowledge	&0.90	&0.88	&0.89\\
        Template 1	&0.89	&0.84	&0.86\\
        Template 2	&0.90	&0.82	&0.85\\
        Template 3	&0.88&	0.84	&0.86\\
        Verbalizer 1	&0.92	&0.91	&0.91\\
        Verbalizer 2	&0.91	&0.90	&0.90\\
        Verbalizer 3	&0.90	&0.87	&0.89\\
        \hline
    \end{tabular}
\end{table}

\subsection{Model Interpretability Study}
This section aims to analyze the interpretability of the model. Unlike general text classification tasks, in disease diagnosis tasks, users are additionally concerned with the interpretability of the results. An unexplainable prediction result is unacceptable to users. Therefore, we conduct an analysis of the model's interpretability, using clinical case examples from Table X.It can be observed that during the knowledge retrieval phase, the following knowledge was extracted from the clinical case: ``Type 2 diabetes is a common chronic disease," ``its main feature is persistently elevated blood sugar levels," ``polydipsia, polyphagia, and polyuria are common symptoms," and ``high blood sugar can lead to kidney disease." From this knowledge, we can identify the key pieces of information from the clinical case that are related to the final prediction result. In other words, we can clearly understand which pieces of knowledge led the model to make the current prediction.

\section{Conclusion}
In this paper, we propose a knowledge-enhanced disease diagnosis method based on a prompt learning framework. This method leverages an external knowledge graph to retrieve relevant knowledge from clinical cases and then encodes this structured knowledge into prompt templates. By incorporating this encoded knowledge, the language model's understanding of the task is improved, resulting in more accurate disease diagnosis outcomes. Experimental results demonstrate that the proposed method effectively enhances the performance of language models in disease diagnosis tasks. Additionally, the model exhibits strong interpretability, providing users with supporting evidence related to the diagnostic results.

In the future, we will explore additional methods for knowledge injection. Additionally, we plan to investigate more advanced knowledge editing techniques to integrate medical knowledge into the reasoning process of language models.

\nocite{*}
\bibliographystyle{plain}
\bibliography{\jobname}

\begin{thebibliography}{10}

\bibitem{ref6}
K.M. Aamir, L.~Sarfraz, M.~Ramzan, et~al.
\newblock A fuzzy rule-based system for classification of diabetes.
\newblock {\em Sensors}, 21(23):8095, 2021.

\bibitem{ref12}
P.~Branco, L.~Torgo, and R.P. Ribeiro.
\newblock A survey of predictive modeling on imbalanced domains.
\newblock {\em ACM Computing Surveys}, 49(2):31, 2016.

\bibitem{ref14}
X.R. Ge, R.D. Williams, and J.A. Stankovic.
\newblock Dkec: domain knowledge enhanced multi-label classification for electronic health record.
\newblock {\em arXiv preprint arXiv:2310.07059}, 2023.

\bibitem{ref9}
S.~Ghimire and H.N. Dhungana.
\newblock Statistical modeling for the prediction of infectious disease dissemination with special reference to covid-19 spread.
\newblock {\em Front Public Health}, 9:1--3, 2021.

\bibitem{ref11}
H.~He and E.A. Garcia.
\newblock Learning from imbalanced data.
\newblock {\em IEEE Transactions on Knowledge and Data Engineering}, 21(9):1263--1284, 2009.

\bibitem{ref4}
M.~Hu, J.~Yang, Y.~Yang, X.~Chen, Y.~Sun, X.~Shen, X.~Wang, T.~Yu, Y.~Mei, L.~Xiao, and W.~Cheng.
\newblock Disease prediction model based on dynamic sampling and transfer learning.
\newblock {\em Journal of Computer Research and Development}, 42(10):2339--2354, 2019.

\bibitem{ref20}
W.X. Liao, B.~Zeng, et~al.
\newblock An improved aspect-category sentiment analysis model for text sentiment analysis based on roberta.
\newblock {\em Applied Intelligence}, 51(6):3522--3533, 2021.

\bibitem{ref2}
F.~Luo, Y.~Zhang, and X.L. Wang.
\newblock Imas++: an intelligent medical analysis system enhanced with deep graph neural networks.
\newblock In {\em Proceedings of the 30th ACM International Conference on Information \& Knowledge Management}, pages 4754--4758, USA, 2021. ACM Press.

\bibitem{ref13}
F.~Luo, Y.~Zhang, and X.L. Wang.
\newblock Imas++: an intelligent medical analysis system enhanced with deep graph neural networks.
\newblock In {\em Proceedings of the 30th ACM International Conference on Information \& Knowledge Management}, pages 4754--4758, USA, 2021. ACM Press.

\bibitem{ref8}
X.~Lü, Q.~Chen, Z.~Zhao, et~al.
\newblock Research on sleep quality evaluation and disease diagnosis based on statistical classification models.
\newblock {\em Value Engineering}, 37(22):273--275, 2018.

\bibitem{ref10}
C.~Ma, C.G. Pan, Z.~Ye, et~al.
\newblock Gout staging diagnosis method based on deep reinforcement learning.
\newblock {\em Processes}, 11(8):2450, 2023.

\bibitem{ref3}
Aytuğ Onan.
\newblock Hierarchical graph-based text classification framework with contextual node embedding and bert-based dynamic fusion.
\newblock {\em J. King Saud Univ. Comput. Inf. Sci.}, 35:101610, 2023.

\bibitem{ref16}
Aytuğ Onan.
\newblock Hierarchical graph-based text classification framework with contextual node embedding and bert-based dynamic fusion.
\newblock {\em J. King Saud Univ. Comput. Inf. Sci.}, 35:101610, 2023.

\bibitem{ref1}
C.~Qu and J.~Bi.
\newblock Application of association rules based on apriori algorithm in disease diagnosis.
\newblock {\em Information and Computer (Theoretical Edition)}, 2015(16):8--9+11, 2015.

\bibitem{ref19}
C.~Raffel, N.~Shazeer, et~al.
\newblock Exploring the limits of transfer learning with a unified text-to-text transformer.
\newblock {\em Journal of Machine Learning Research}, 21(2):1--14, 2020.

\bibitem{ref7}
J.A. Sanz, M.~Galar, A.~Jurio, et~al.
\newblock Medical diagnosis of cardiovascular diseases using an interval-valued fuzzy rule-based classification system.
\newblock {\em Applied Soft Computing}, 20:103--111, 2014.

\bibitem{ref18}
Shyni Sharaf and V.S. Anoop.
\newblock An analysis on large language models in healthcare: A case study of biobert.
\newblock {\em ArXiv}, abs/2310.07282:n. pag., 2023.

\bibitem{ref5}
E.H. Shortliffe.
\newblock Computer-based medical consultations: Mycin.
\newblock {\em Elsevier}, 85(6):243--260, 1976.

\bibitem{ref15}
H.~Tian and C.~Xu.
\newblock Research on medical short text classification algorithm based on bert model.
\newblock {\em Journal of Ili Normal University (Natural Science Edition)}, 15(04):50--57, 2021.

\bibitem{ref17}
L.~Xu, D.~Li, H.~Zhang, et~al.
\newblock Research on medical text classification based on neural networks.
\newblock {\em Computer Engineering and Science}, 45(06):1116--1122, 2023.

\end{thebibliography}

\end{document}